\documentclass[10pt,twocolumn,letterpaper]{article}

\usepackage{cvpr}              
\definecolor{cvprblue}{rgb}{0.21,0.49,0.74}
\usepackage[pagebackref,breaklinks,colorlinks,allcolors=cvprblue]{hyperref}
\usepackage{bm}
\usepackage{pifont}
\usepackage{multirow}
\usepackage[table]{xcolor}  
\usepackage{tcolorbox}
\usepackage[accsupp]{axessibility} 

\def\paperID{} 
\def\confName{CVPR}
\def\confYear{2026}

\title{DRM: Diffusion-based Reward Model With Step-wise Guidance}

\author{
Jaxon Zhang$^{1*}$,
Binxin Yang$^{2}$,
Hubery Yin$^{2}$,
Chen Li$^{2}$,
Jing LYU$^{2}$\\
$^1$Peking University \quad $^2$WeChat Vision, Tencent Inc.\\
}

\begin{document}
\maketitle
\begin{abstract}
Current mainstream methods of aligning diffusion models with human preferences typically employ VLM-based reward models.
However, these reward models, pre-trained for semantic alignment, struggle to capture the essential perceptual qualities—such as aesthetics, composition, and visual harmony.
In this work, we argue that a model capable of high-fidelity generation must possess a profound understanding of these visual attributes.
Based on this insight, we introduce the Diffusion-based Reward Model (DRM), a novel paradigm that use the pre-trained diffusion model as a powerful evaluative backbone.
A key advantage of the DRM is its unique ability to assess not only the final image but also the noisy intermediate latents at any stage of the generative process. 
We leverage this step-wise evaluative capacity in two ways.
First, we propose Step-wise GRPO, a reinforcement learning algorithm that provides dense, per-step rewards to resolve the imprecise credit assignment problem in GRPO algorithm, leading to more stable and effective alignment.
Second, we introduce Step-wise Sampling, a novel inference strategy that employs the DRM as a dynamic guide to evaluate multiple generation paths at each step, steering the process towards higher-quality outcomes.
Extensive experiments confirm that our approach significantly enhances the final quality of generated images. 
Code: \url{https://github.com/jjaxonx/DRM}.
\end{abstract}    
\section{Introduction}
\label{sec:intro}

\renewcommand{\thefootnote}{\fnsymbol{footnote}}
\footnotetext[1]{Work done during an internship at WeChat Vision, Tencent Inc.}

\begin{figure}[t]
  \centering
    \includegraphics[width=0.925\linewidth]{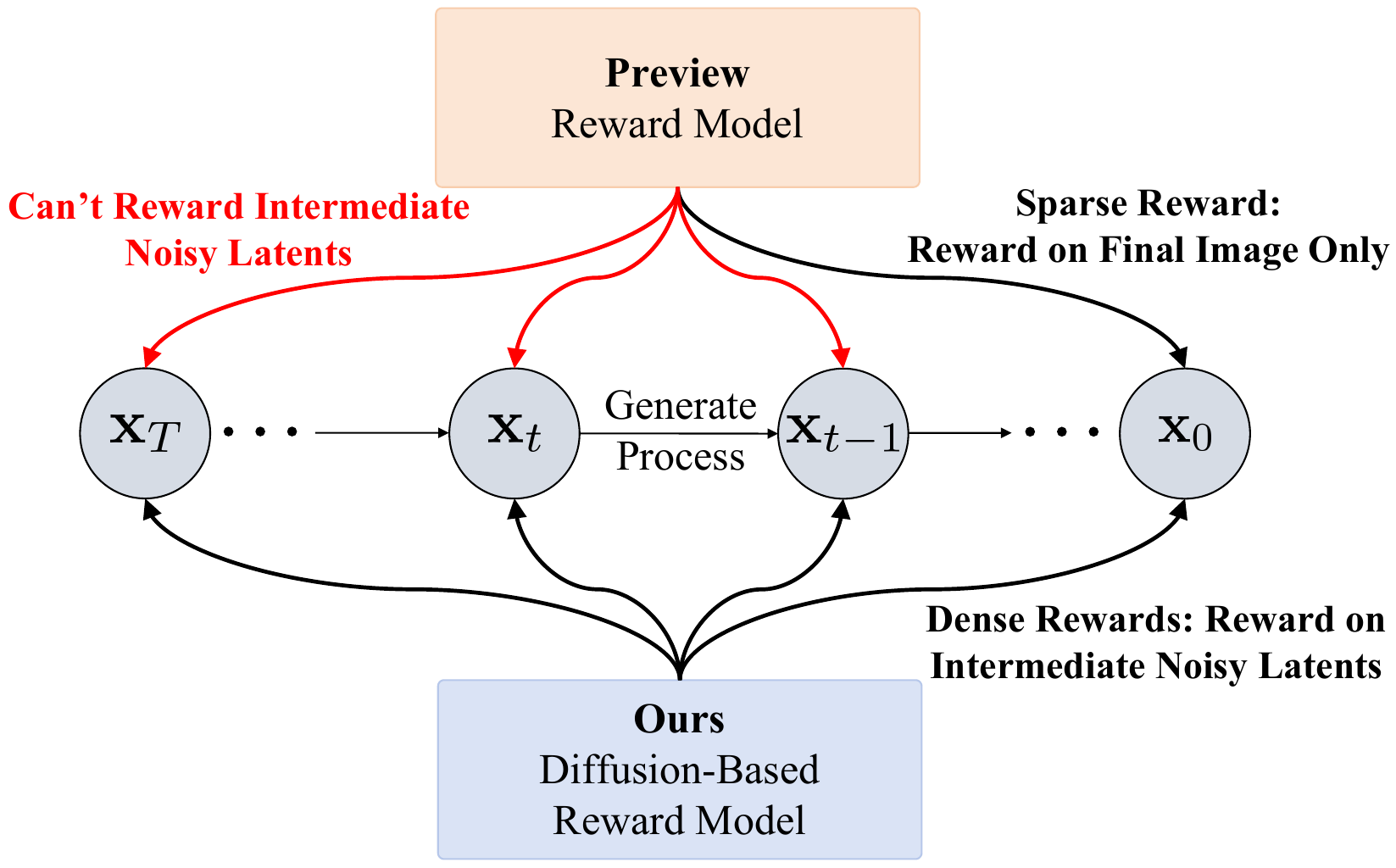}
    \caption{
    \textbf{Comparison between preview reward models and DRM.} 
    Existing reward models treat the generation process as a black box, providing only a single, terminal reward based on the final output. 
    Our DRM offers fine-grained reward for any noisy latent along the entire denoising trajectory.}
    \vspace{-2.5mm}
\label{fig:teaser}
\end{figure}

\begin{figure}[t]
  \centering
    \includegraphics[width=0.90\linewidth]{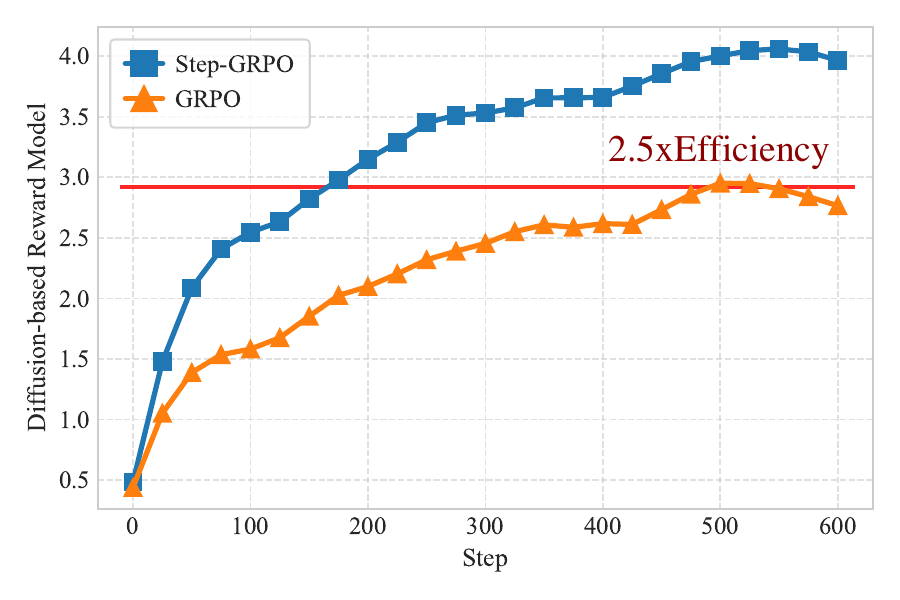}
    \vspace{-4mm}
    \caption{
    \textbf{Reward curves for various RL algorithms optimized using our DRM.}
    Our Step-GRPO, which leverages dense, per-step rewards, not only reaches a higher final reward but also converges \textbf{2.5x faster on step} than the standard GRPO baseline.
    }
    \vspace{-5mm}
\label{fig:reward_line}
\end{figure}

Diffusion models~\citep{ho2020denoising, saharia2022photorealistic, rombach2022high, podell2023sdxl, esser2024scaling} have demonstrated remarkable generative capabilities.
However, their outputs often misalign with human preferences and intent, spurring the wave of research into human preference alignment.
A direct approach to alignment involves fine-tuning the model on large-scale human feedback~\cite{liu2025videodpo, wallace2024diffusion}.
While effective, this process is prohibitively expensive and labor-intensive for diffusion models.
Consequently, an alternative paradigm has gained prominence: learning a reward model (RM) from limited preference data~\cite{lee2023aligning, liu2025improving, wang2024lift, xu2024visionreward, zang2025simple}.
This learned reward function can then be used to generate synthetic preference data, significantly reducing the reliance on manual annotation for model alignment.

Early RMs~\cite{liang2024rich, xu2023imagereward, zhang2024learning} were typically fine-tuned from CLIP~\cite{radford2021learning} backbones.
With the advent of Vision-Language Models (VLMs)~\cite{achiam2023gpt, liu2023visual, wang2024qwen2, bai2025qwen2}, their superior visual understanding capabilities made them a more powerful choice for RM backbones, leading to their widespread adoption in visual quality assessment~\cite{he2024videoscore, liu2025improving, wang2024lift, xu2024visionreward, wang2025unified, li2025q, zhang2025vq, cao2025artimuse}.
However, these VLMs rely on a CLIP-style vision encoder, which is pre-trained to align images with text based on semantic similarity.
This objective inherently prioritizes what an image contains over how it is presented.
Consequently, the resulting feature representations are rich in semantics but impoverished in terms of crucial aesthetic and compositional attributes that are pivotal to human preference.

This limitation motivates the search for an alternative vision backbone, one inherently sensitive to the perceptual qualities that CLIP-style encoders neglect.
We argue that pre-trained diffusion models are precisely such a backbone. 
This claim is built on an intuitive yet powerful insight: the ability to generate high-fidelity images necessitates a deep, implicit understanding of visual aesthetics, composition, and fine-grained details.
Motivated by this insight, we pioneer the use of diffusion models as the backbone for reward modeling, systematically unlocking their powerful evaluative capabilities.
We introduce our approach as the \textbf{Diffusion-based Reward Model (DRM)}.

The benefit of using a diffusion backbone is clear: a richer understanding of perceptual qualities like aesthetics and composition.
Beyond this, the DRM possesses a more profound advantage:
as shown in Figure~\ref{fig:teaser}, it does not merely judge the final image; it comprehends the entire generative trajectory, allowing it to assess noisy intermediate states at any given timestep.
This unique, step-wise evaluative capacity provides a mechanism to address two key challenges in diffusion models.

\textbf{(1)}
On the optimization front, prevailing reinforcement learning alignment methods, such as GRPO~\cite{liu2025flow, xue2025dancegrpo}, suffer from an imprecise credit assignment problem.
They treat the multi-step generation process as a ``black box," uniformly distributing the reward from the final image across all intermediate timesteps.
This coarse approach fails to distinguish between beneficial and detrimental actions during generation.
To resolve this, our Step-wise GRPO (Step-GRPO) algorithm leverages the DRM to provide immediate and precise rewards at each step.
As visualized in Figure~\ref{fig:reward_line}, this dense feedback signal enables far more effective and stable policy optimization.
\textbf{(2)}
For inference, we break the rigidity of deterministic samplers.
Where conventional methods are locked into a single, uncorrectable path, our Step-wise Sampling strategy employs the DRM as a dynamic guide.
At each step, it evaluates multiple potential futures and greedily chooses the one that best preserves quality, preventing the cascading failures common in fixed trajectories.
In summary, our contributions are as follows:

\noindent \ding{113}~(1) 
We introduce the DRM, a novel paradigm for reward modeling. 
By using a pre-trained diffusion model as its backbone, the DRM inherits a rich understanding of perceptual qualities like aesthetics and composition, and crucially, it possesses the unique capability to assess noisy intermediate latents at any stage of the generative process.

\noindent \ding{113}~(2) 
We propose Step-GRPO, a reinforcement learning algorithm that resolves the credit assignment problem in diffusion model alignment. 
By using the DRM to provide dense, per-step rewards, Step-GRPO achieves significantly more stable and efficient policy optimization compared to methods that rely on a single, terminal reward.

\noindent \ding{113}~(3)
We present Step-wise Sampling, a novel inference strategy that employs the DRM as a dynamic guide.
This method evaluates multiple potential generation paths at each step, steering the process towards higher quality outcomes.


\section{Related Work}
\label{sec:formatting}

\subsection{Reward Model}
Reward models are crucial for aligning diffusion generative models~\citep{ho2020denoising, nichol2021glide, saharia2022photorealistic, rombach2022high, podell2023sdxl, esser2024scaling, flux, zhang2026alignedgen, lv2025rethinking, jiang2026octot2i} with human preferences. 
Initially, methods relied on automated metrics like FID~\cite{heusel2017gans} and CLIP~\cite{radford2021learning} to evaluate image quality and text-image consistency~\cite{huang2023t2i, huang2024vbench, liu2024evalcrafter}. 
However, these metrics fall short of capturing human preferences due to training objectives and data. 
To bridge this gap, recent research focuses on fine-tuning CLIP models directly on human preference datasets, enabling them to better predict human judgments~\cite{liang2024rich, xu2023imagereward, zhang2024learning}. 
With the rise of powerful Vision Language Models (VLMs)~\cite{achiam2023gpt, liu2023visual, wang2024qwen2, bai2025qwen2}, they have become a natural choice for reward model backbones, leading to their widespread adoption in visual quality assessment~\cite{he2024videoscore, liu2025improving, wang2024lift, xu2024visionreward, wang2025unified, li2025q, zhang2025vq}. 
A key limitation, however, is that these VLMs use a CLIP-style vision encoder that compresses an image into a semantic-heavy representation. 
This information bottleneck makes the VLM less sensitive to the image's structural integrity and other details.
While LPO~\cite{zhang2025diffusion} have explored diffusion-based reward models, these efforts lack a systematic investigation.
Motivated by the premise that ``generation requires understanding," we systematically explore the diffusion model as a reward backbone, aiming to unlock its potential for more perceptive and accurate reward signals.

\subsection{Alignment for Diffusion Models}
Aligning diffusion models with human preferences is a significant area of investigation. 
Recent efforts have largely followed two paths. 
One line of work adapts direct preference optimization (DPO)~\cite{rafailov2023direct} for diffusion models, as seen in D3PO~\cite{yang2024using} and Diffusion-DPO~\cite{wallace2024diffusion}. The other integrates online reinforcement learning, with Flow-GRPO~\cite{liu2025flow} and DanceGRPO~\cite{xue2025dancegrpo} being the first to apply it to flow-matching models, inspiring a surge of subsequent works~\cite{liu2025diversegrpo, wang2025grpo, li2025mixgrpo, li2025branchgrpo, jin2026dgpo, lin2026visd}.
A central challenge in these approaches is the problem of credit assignment: the reward for the final image is uniformly applied to all steps in the generation process. 
TempFlow-GRPO~\cite{he2025tempflow} attempts to solve this with a precise score allocation mechanism, but this introduces substantial sampling overhead during training. 
Our DRM is designed to address this challenge directly. By possessing the inherent capability to evaluate intermediate noisy latents, it offers a more direct and efficient approach to the credit assignment problem.

\section{Method}

\begin{figure*}[!t]
\vspace{0pt}
\centering
\includegraphics[width=0.95\textwidth]{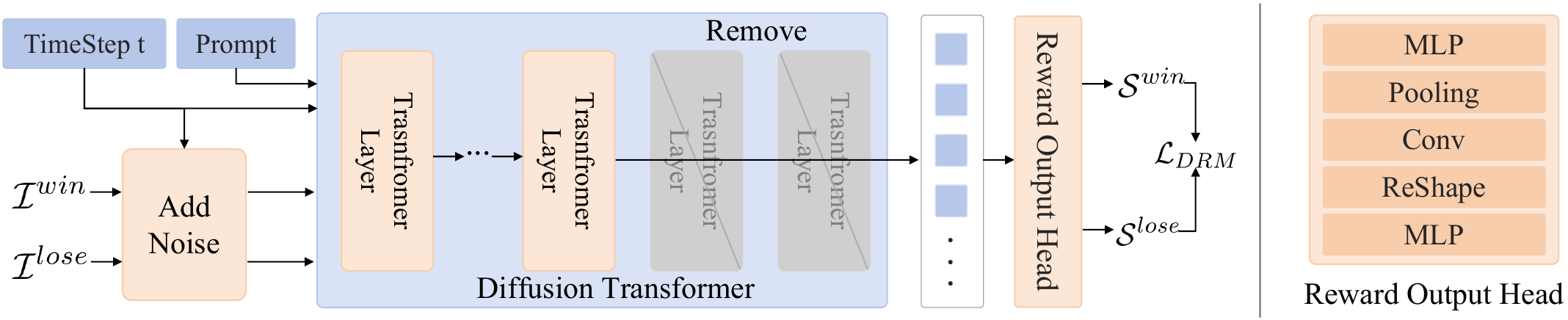}
\vspace{-5pt}
\caption{
\textbf{Overview of the Diffusion-based Reward Model (DRM).} (Left) The training pipeline. During training, the DRM takes a pair of preferred and dispreferred images, both corrupted with noise at a specific timestep $t$, and predicts their respective reward scores. The model is then optimized via DRM loss. (Right) The detailed architecture of our Reward Output Head.
}
\label{fig:arch}
\vspace{-2.5pt}
\end{figure*}

\begin{figure*}[!t]
\vspace{0pt}
\centering
\includegraphics[width=0.95\textwidth]{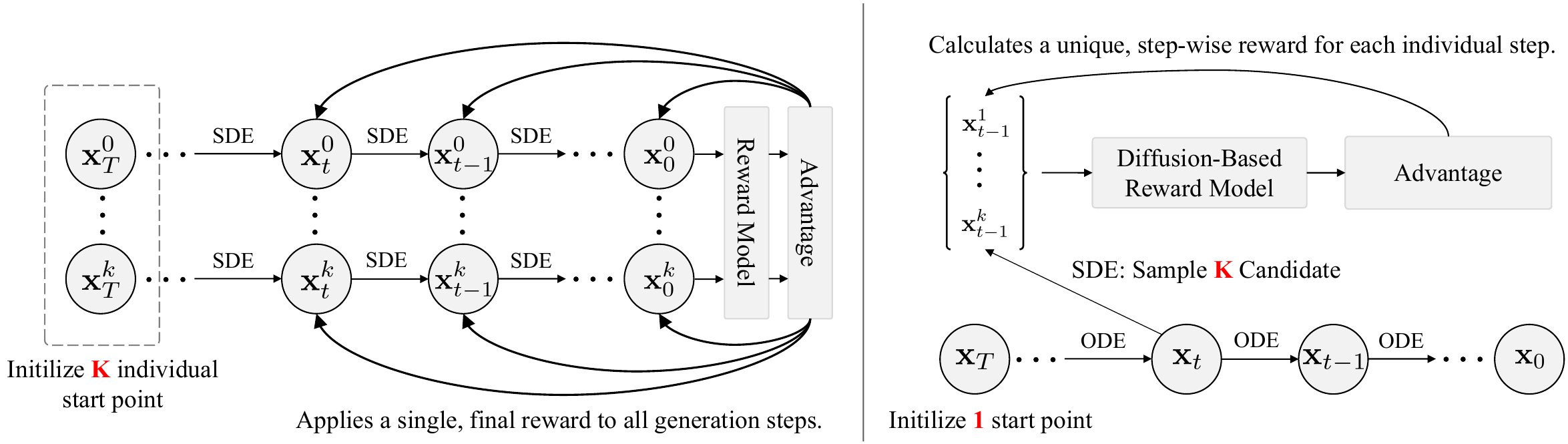}
\vspace{-2.5pt}
\caption{
\textbf{GRPO vs. Step-wise GRPO.}
(Left) Naive GRPO relies on a terminal reward. 
It samples multiple full trajectories, calculates a single reward at the final step (t=0), and applies this coarse reward uniformly to all preceding steps, leading to imprecise credit assignment. 
(Right) Step-wise GRPO introduces a dense, per-step reward signal.
From a single initial point, it explores k candidate samples via SDE at each timestep, using the DRM to assign a precise, step-specific reward and advantage for more effective policy optimization.
}
\label{fig:stepgrpo}
\vspace{-2.5pt}
\end{figure*}

\subsection{Preliminary}

\paragraph{Flow Matching.} 
Let $x_0 \sim \mathcal{X}_0$ be a data sample from the real world image data distribution and $x_1 \sim \mathcal{X}_1$ a noise sample. Flow Matching\citep{lipman2022flow} defines intermediate samples as
\begin{equation}
x_t = (1-t)x_0 + t x_1, \quad t \in [0,1],
\end{equation}
and trains a velocity field $v_\theta(x_t,t)$ via the objective:
\begin{equation}
\mathcal{L}_{\text{FM}}(\theta) = \mathbb{E}_{t,x_0,x_1} \big[\|v - v_\theta(x_t,t)\|_2^2 \big], \quad v = x_1 - x_0.
\end{equation}
At inference, the iterative denoising process can be naturally formalized as a Markov Decision Process \citep{black2023training}. At each step $t$, the state is $s_t = (c, t, x_t)$, where $c$ denotes the prompt, and the action $a_t$ corresponds to producing the denoised sample $x_{t-1} \sim \pi_\theta(x_{t-1}|x_t,c)$.

\paragraph{Flow-GRPO.} 
RL aims to learn a policy that maximizes the expected cumulative reward. 
Given a prompt $\mathbf{c}$, the flow model $p_{\theta}$ samples a group of $G$ individual images $\{\mathbf{x}_0^i\}_{i=1}^G$ and the corresponding reverse-time trajectories $\{(\mathbf{x}_T^i, \mathbf{x}_{T-1}^i, ..., \mathbf{x}_0^i)\}_{i=1}^G$. Then, the advantage of the $i$-th image is calculated by normalizing the group-level rewards as follows:
\begin{equation}
\label{eq:advantage}
    \hat{A}_t^i = \frac{R(\mathbf{x}_0^i, \mathbf{c}) - \text{mean}(\{R(\mathbf{x}^i_0,\mathbf{c})\}_{i=1}^G)}{\text{std}(\{R(\mathbf{x}_0^i,\mathbf{c})\}_{i=1}^G)}.
\end{equation}
GRPO optimizes the policy model by maximizing the following objective:
\begin{equation}
\label{eq:grpo}
    \mathcal{J}_{\text{Flow-GRPO}}(\theta) = \mathbb{E}_{\mathbf{c}\sim\mathcal{C}, \{\mathbf{x}^i\}_{i=1}^G\sim\pi_{\theta_{\text{old}}}(\cdot|\mathbf{c})}f(r, \hat{A}, \theta, \epsilon, \beta),
\end{equation}
where
\begin{equation}
\begin{split}
    f(r, \hat{A}, \theta, \epsilon, \beta) = \frac{1}{G} \sum_{i=1}^{G} \frac{1}{T} \sum_{t=0}^{T-1} \Bigl( \min\bigl(r_t^i(\theta)\hat{A}_t^i,\, \\ \operatorname{clip}(r_t^i(\theta), 1-\epsilon, 1+\epsilon)\hat{A}_t^i \bigr) 
     - \beta D_{\mathrm{KL}}(\pi_{\theta} \,&\|\,\pi_{\text{ref}}) \Bigr), 
\end{split}
\end{equation}
with $r_t^i(\theta) = \frac{p_\theta(x_{t-1}^i|x_t^i,c)}{p_{\theta_{\text{old}}}(x_{t-1}^i|x_t^i,c)}$. 
To satisfy GRPO’s stochastic exploration requirements, \citep{liu2025flow} convert the deterministic ODE to an equivalent SDE:
\begin{flalign}
\begin{split}
  x_{t+\Delta t} = x_t + {} \Bigl(v_\theta(x_t,t) + \\ \frac{\sigma_t^2}{2t}\bigl(x_t + (1-t)&v_\theta(x_t,t)\bigr)\Bigr)\Delta t  + \sigma_t \sqrt{\Delta t} \,\epsilon,
\end{split}
\end{flalign}
where $\mathbf{\epsilon} \sim \mathcal{N}(0,\mathbf{I})$ injects stochasticity and $\sigma_t = a\sqrt{\frac{t}{1-t}}$.

\subsection{Diffusion-based Reward Model}
\paragraph{Architecture of Diffusion-based Reward Model (DRM).} 
As illustrated in Figure~\ref{fig:arch}, our DRM predicts human preferences by leveraging the intermediate features from the Diffusion Transformer (DiT) of a pre-trained diffusion model.
Specifically, to ensure a fair comparison with VLM-based reward models in terms of parameter count, we adapt a pre-trained DiT by truncating its final transformer layers.
For instance, we initialize our backbone with the pre-trained DiT from SD3.5-Medium (2.5B parameters).
To align its scale with models like HPSv3-2B, we remove the last three transformer layers.
Given a noisy latent representation $x_{t}$ at a specific timestep $t$, it is fed into our modified DiT backbone. 
This process yields a sequence of visual features $f_{v} \in \mathbb{R}^{L \times d}$, where $L$ is the sequence length and $d$ is the feature dimension. 
The visual features $f_{v}$ are then passed to a prediction head. First, a linear layer projects them to a lower-dimensional space, resulting in $f_{p} \in \mathbb{R}^{L \times d_{p}}$. Subsequently, $f_{p}$ is reshaped into a spatial feature map $f_{p} \in \mathbb{R}^{h \times w \times \frac{d_{p}}{4}}$. 
Finally, this feature map is processed by a small convolutional network, followed by a pooling layer and a liner projection, to produce the final preference score $s$. 
The overall process of DRM can be formulated as follows:
\begin{equation}
\begin{aligned}
\label{eq:arch1}
f_{p} &= \text{MLP}(f_{v}),\ f_v \in \mathbb{R}^{L \times d}, f_{p} \in \mathbb{R}^{L \times d_{p}} \ \\
\end{aligned}
\end{equation}
\begin{equation}
\begin{aligned}
\label{eq:arch2}
s &= \text{MLP}(\text{Pooling}(\text{Conv}(\text{ReShape}(f_{p}))))
\end{aligned}
\end{equation}

\paragraph{Training Loss.} 
Our model is trained on a dataset composed of triplets $(I^{win}, I^{lose}, p)$, where $(I^{win}, I^{lose})$ represents a pair of images with human preference labels (winner and loser), and $p$ is their corresponding text prompt.
The training process for a given pair is as follows, also illustrated in Figure~\ref{fig:arch}. First, we encode the images into latent representations, $x_0^{win}$ and $x_{0}^{lose}$, using a VAE encoder. 
Subsequently, for a randomly sampled timestep $t$, we simulate the forward diffusion process by adding Gaussian noise $\epsilon_{t} \in \mathcal{N}(0, 1)$ to generate the noisy latents $x_{t}^{win}$ and $x_{t}^{lose}$.
These noisy latents are then fed into our DRM, conditioned on the timestep $t$, to obtain their respective preference scores:
\begin{equation}
\begin{aligned}
\label{eq:score}
s^{win} = \text{DRM}(x^{win}_{t}, t),\ s^{lose} = \text{DRM}(x^{lose}_{t}, t)
\end{aligned}
\end{equation}
Following the Bradley-Terry (BT) model~\cite{bradley1952rank}, we define the training loss as the negative log-likelihood of the probability that the winning image is preferred over the losing one:
\begin{equation}
\begin{aligned}
\label{eq:loss}
\mathcal{L}_{DRM}=-\text{log}(\sigma(s^{win}-s^{lose})),
\end{aligned}
\end{equation}
where $\sigma(\cdot)$ denotes the sigmoid function.

\subsection{Step-wise GRPO}
\paragraph{Motivation.}
Prevailing reinforcement learning (RL) alignment algorithms, such as GRPO, largely treat the multi-step generation process as a ``black box", performing time-agnostic policy optimization.
This approach suffers from a fundamental limitation: it assigns the reward signal from the final generated image uniformly to every step in the generation trajectory. 
This coarse credit assignment mechanism overlooks the varying contributions of each intermediate step to the final image quality.
To address this core issue, we leverage a unique capability of our Diffusion-based Reward Model (DRM). 
Because its backbone is initialized from pre-trained diffusion model weights, the DRM is inherently capable of evaluating noisy latents at any arbitrary timestep during the generation process. 
Capitalizing on this property, we introduce the Step-wise GRPO (Step-GRPO) algorithm. 
Instead of relying on a single, terminal reward, Step-GRPO provides a precise, step-specific reward for each intermediate state, enabling a more granular and effective policy optimization.

\paragraph{Step-wise GRPO (Step-GRPO).}
As illustrated in Figure~\ref{fig:stepgrpo}, our method performs fine-grained policy optimization at each reverse diffusion timestep $t$.
Specifically, starting from the current state $\mathbf{x}_{t+1}$, we sample a set of $k$ candidate states for the next step, $\{\mathbf{x}_t^{i}\}_{i=1}^k$, via the SDE.
These candidates are then fed into our DRM to obtain a corresponding set of immediate reward scores $\{R(\mathbf{x}_t^{i}, c)\}_{i=1}^k$
Unlike conventional advantage functions (Equal ~\ref{eq:advantage}) that rely on a terminal reward, we define an immediate advantage for the decision at each step, formulated as:
\begin{equation}
\label{eq:stepadvantage}
    \hat{A}_{t}^i = \frac{R(\mathbf{x}_t^i, \mathbf{c}) - \text{mean}(\{R(\mathbf{x}_t^i,\mathbf{c})\}_{i=1}^k)}{\text{std}(\{R(\mathbf{x}_t^i,\mathbf{c})\}_{i=1}^k)}.
\end{equation}
This formulation shifts the focus of evaluation from the final, global outcome to the local decision at the current timestep—specifically, assessing the relative quality of transitioning from $\mathbf{x}_t$ to each candidate $\mathbf{x}_t^{i}$. 
This approach yields a more precise advantage estimate and provides a more direct and fine-grained supervisory signal for the policy gradient.

\begin{figure}[t]
  \centering
    \includegraphics[width=0.825\linewidth]{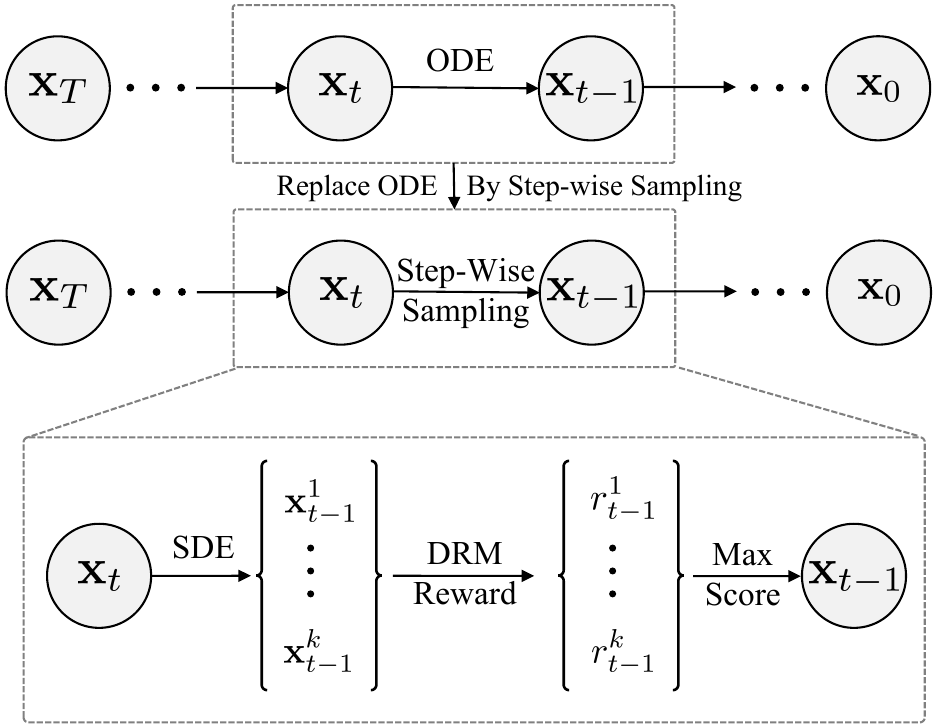}
    \vspace{-2.5pt}
    \caption{\textbf{Overview of Step-wise Sampling.}
    At each step t, we perform a branching into k candidates via SDE. The DRM scores these candidates, and the top-scoring latent is chosen to continue the trajectory.}
    \vspace{-15pt}
\label{fig:stepsample}
\end{figure}

\subsection{Step-wise Sampling}
Beyond its role in providing step-wise rewards for RL-based fine-tuning, our Diffusion-based Reward Model (DRM) can also be leveraged to directly enhance generation quality at inference time. 
We introduce a novel sampling strategy, termed Step-wise Sampling, which offers a training-free, plug-and-play mechanism for improving model outputs. 
This approach provides a highly practical method for users to boost generation quality without any model fine-tuning.
Conventional deterministic samplers follow a single, fixed trajectory, generating only one successor state $\mathbf{x}_{t-1}$ from $\mathbf{x}_t$ at each timestep. 
While efficient, this single-path approach is unforgiving; the quality of the final image is entirely dependent on the model's initial predictions, with no opportunity for corrective action during the generation process.

To overcome this limitation, Step-wise Sampling introduces an ``explore-and-select" mechanism, as illustrated in Figure ~\ref{fig:stepsample}.
At each timestep t, instead of following a single deterministic path, we first ``explore" by leveraging SDE sample to generate k candidate states for the next step, forming a candidate set $\{\mathbf{x}_{t-1}^i\}_{i=1}^k$. 
This step effectively branches the generation process into multiple potential future trajectories.
Next, in the ``select" phase, we harness the unique power of our DRM to score each of these k candidates, obtaining a set of corresponding rewards $\{R(x_{t-1}^i, c)\}_{i=1}^k$. 
We greedily select the candidate with the highest score as the definitive state for the next step:
\begin{equation}
\label{eq:stepsample}
    \mathbf{x}_{t-1} = \text{argmax}_{\mathbf{x}_{t-1}^i} ( R(\mathbf{x}_{t-1}^i, c)).
\end{equation}
By iteratively selecting the most promising path at each stage of generation, Step-wise Sampling can proactively steer the trajectory away from "bad" paths that might lead to low-quality results.
This process robustly enhances the quality and alignment of the final image. 

\section{Experiment}

\subsection{Diffusion-based Reward Model}

\subsubsection{Implementation Details}
\paragraph{Training Dataset.}
Following the methodology of HPSv3, we construct our training dataset by aggregating data from three sources: HPDv3, a subset of the Pick-A-Pic dataset, and a subset of the ImageReward dataset. 
The final dataset comprises a total of 1.4 million samples. 
Each sample is structured as a triplet $(I^{win}, \ I^{lose}, p)$, consisting of a preferred image $I^{win}$, a dispreferred image $I^{lose}$, and their shared text prompt $p$.
\paragraph{Model.}
We initialize our backbone with the pretrained Diffusion Transformer from SD3.5-Medium. 
To ensure a fair comparison with VLM-based models in terms of parameter count, we truncate the final three transformer layers of the model. 
All remaining parameters are made trainable and are fine-tuned during training.
The model is trained for one epoch on a cluster of 64 NVIDIA H20 GPUs, each with 96 GB of VRAM. 
We employ a constant learning rate of $1\times 10^{-5}$, and a global batch size of 128, which corresponds to a per-GPU batch size of 2. 
All images are resized to a resolution of 512×512 pixels at training.

\begin{table*}[!t]
\centering
\footnotesize
\setlength{\tabcolsep}{2.5mm}{
\begin{tabular}
{c|c|c|c|c|cccc}
\toprule
\multicolumn{1}{c}{\#} & \multicolumn{1}{|c}{Model} & \multicolumn{1}{|c}{Weights} & \multicolumn{1}{|c}{Epoch} & \multicolumn{1}{|c}{Size} & \multicolumn{1}{|c}{$\textbf{ImageReward}$ $\uparrow$} & \multicolumn{1}{c}{$\textbf{PickScore}$ $\uparrow$} & \multicolumn{1}{c}{$\textbf{HPDv2}$ $\uparrow$} & \multicolumn{1}{c}{$\textbf{HPDv3}$ $\uparrow$} \\
\midrule
- & {CLIP ViT-H/14~\cite{radford2021learning}} & - & - & - & 57.1 & 60.8 & 65.1 & 48.6 \\
- & {Aesthetic Score Predictor~\cite{schuhmann2022laion}} & - & - & - & 57.4 & 56.8 & 76.8 & 59.9 \\
- & {ImageReward~\cite{xu2023imagereward}} & - & - & - & 65.1 & 61.1 & 74.0 & 58.6 \\
- & {PickScore~\cite{kirstain2023pick}} & - & - & - & 61.6 & {70.5} & 79.8 & {65.6} \\
- & {HPS~\cite{wu2023better}} & - & - & - &  61.2 & 66.7 & 77.6 & 63.8 \\
- & {HPSv2~\cite{wu2023human}} & - & - & - & 65.7 & 63.8 & {83.3} & 65.3 \\
- & {MPS~\cite{zhang2024learning}} & - & - & - & \textbf{67.5} & 63.1 & \underline{83.5} & 64.3 \\
- & {HPSv3 - 2B~\cite{ma2025hpsv3}} & - & - & - & 57.9 &  63.6 &  80.8 &  66.3 \\
- & {HPSv3 - 7B~\cite{ma2025hpsv3}} & - & - & - & \underline{66.8} &  \underline{72.8} &  \textbf{85.4} &  \textbf{76.9} \\
\midrule
(a) & Ours & Random & 1 & 256 &  {52.4} &  {57.5} &  {65.0} &  {59.3} \\
(b) & Ours & Random & 2 & 256 &  {51.9} &  {59.5} &  {68.5} &  {62.3} \\
(c) & Ours & Random & 3 & 256 &  {53.7} &  {59.0} &  {70.1} &  {63.0} \\
\midrule
(d) & Ours & Pre-trained & 1 & 256 &  {62.9} &  {72.1} &  {80.1} &  {71.9} \\
\midrule
\rowcolor{gray!12.5}
(e) & Ours & Pre-trained & 1 & 512 &  {64.1} &  \textbf{73.4} &  {82.2} &  \underline{74.0} \\
\bottomrule
\end{tabular}
}
\vspace{0pt}
\caption{\textbf{Preference prediction accuracy (\%) on the test sets of ImageReward, HPDv2 and HPDv3.} The best and second-best results are \textbf{bolded} and \underline{underlined}.
Our model achieves top-tier accuracy on PickScore.
Its competitive scores on ImageReward, HPDv2 and HPDv3 reflect an expected trade-off, stemming from the DRM's core design.
The DRM is trained to assess noisy latents throughout the generation process, not just the final clean outputs.
This capability, fundamental to our approach, introduces a subtle domain shift when evaluated on benchmarks consisting solely of clean images, which accounts for the performance gap.
}
\label{table:drm}
\vspace{-10pt}
\end{table*}

\subsubsection{Preference Comparison}
We evaluated our model against several leading reward models on standard benchmarks.
As shown in Table~\ref{table:drm}, our approach achieves highly competitive performance, securing accuracies of 64.1\% 73.4\%, 82.2\%, and 74.0\% on the PickScore, HPDv2, and HPDv3 test sets, respectively.
It is crucial to contextualize these results: unlike conventional RMs that are trained exclusively to judge final, clean images, our DRM is designed for the more challenging and general task of evaluating noisy latents at any step of the generation process.
This broader training objective, which is fundamental to enabling our step-wise guidance methods, is not measured by standard benchmarks.
This inherent design may introduce a slight trade-off in performance on clean-only evaluation tasks.
In addition, the key advantage of our model lies in its remarkable parameter efficiency.
Despite its modest size of only 2B parameters, our DRM significantly outperforms the similarly-sized, VLM-based HPSv3. 
This evidence strongly suggests that our diffusion-based architecture provides a more efficient and effective pathway to powerful reward modeling than simply scaling up conventional VLM backbones.

\subsubsection{Ablation Study}
\paragraph{Effect of Pretrained Weight.}
To validate our hypothesis that the model's strong performance stems from the generative prior embedded in the pre-trained diffusion weights, we conducted a critical ablation study. Specifically, we trained an identical model architecture from scratch, using random weight initialization instead of loading the pre-trained weights.
The results after a single epoch of training are stark.
As shown by comparing rows (a) and (e) in Table~\ref{table:drm}, the model initialized with pre-trained weights significantly outperforms the randomly initialized version across all test sets.
To rule out the possibility that this discrepancy was merely due to the from-scratch model not having converged, we extended its training duration.
The subsequent results, presented in rows (b) and (c), are conclusive: the pre-trained diffusion weights not only dramatically accelerate convergence but also enable the model to reach a higher performance ceiling. 
This confirms that the generative prior is indispensable for both training efficiency and the model's ultimate evaluative capabilities.

\paragraph{Effect of Training Image Size.}
To assess the impact of training image resolution on model performance, we conducted an ablation study comparing models trained on 256x256 and 512x512 images.
The results are presented in Table~\ref{table:drm}, comparing rows (d) and (e).
A clear trend emerges: increasing the training resolution leads to a consistent improvement in performance across all test sets.
This indicates that our DRM has the capacity to leverage the fine-grained details present in higher-resolution data to make more accurate judgments, highlighting the importance of high-resolution training for achieving optimal performance.

\begin{table}
  \centering
  \footnotesize
  \setlength\tabcolsep{4.0mm}{
  \scalebox{1.00}{
  \begin{tabular}{cccc}
    \toprule
    \textbf{Timestep} & \textbf{0} & \textbf{500} & \textbf{750} \\
    \midrule
    DRM & 74.0 & 73.0  & 65.11 \\
    \bottomrule
  \end{tabular}
  }}
  \caption{\textbf{DRM Accuracy vs. Timestep.}
  Performance on HPSv3 test set. Higher timestep correspond to higher noise level.}
  \label{tab:noise}
  \vspace{-10pt}
\end{table}

\subsubsection{Influence of Timestep}
To validate the DRM's core capability of evaluating noisy latents, we tested its preference prediction accuracy on the HPSv3 test set across a spectrum of timestep.
The results, detailed in Table~\ref{tab:noise}, show a predictable decline in performance as the signal-to-noise ratio decreases. 
Nevertheless, the DRM's accuracy remains remarkably robust, confirming its efficacy as a reliable, step-wise reward signal throughout the entire denoising trajectory.

\begin{figure*}[!t]
\vspace{0pt}
\centering
\includegraphics[width=0.725\textwidth]{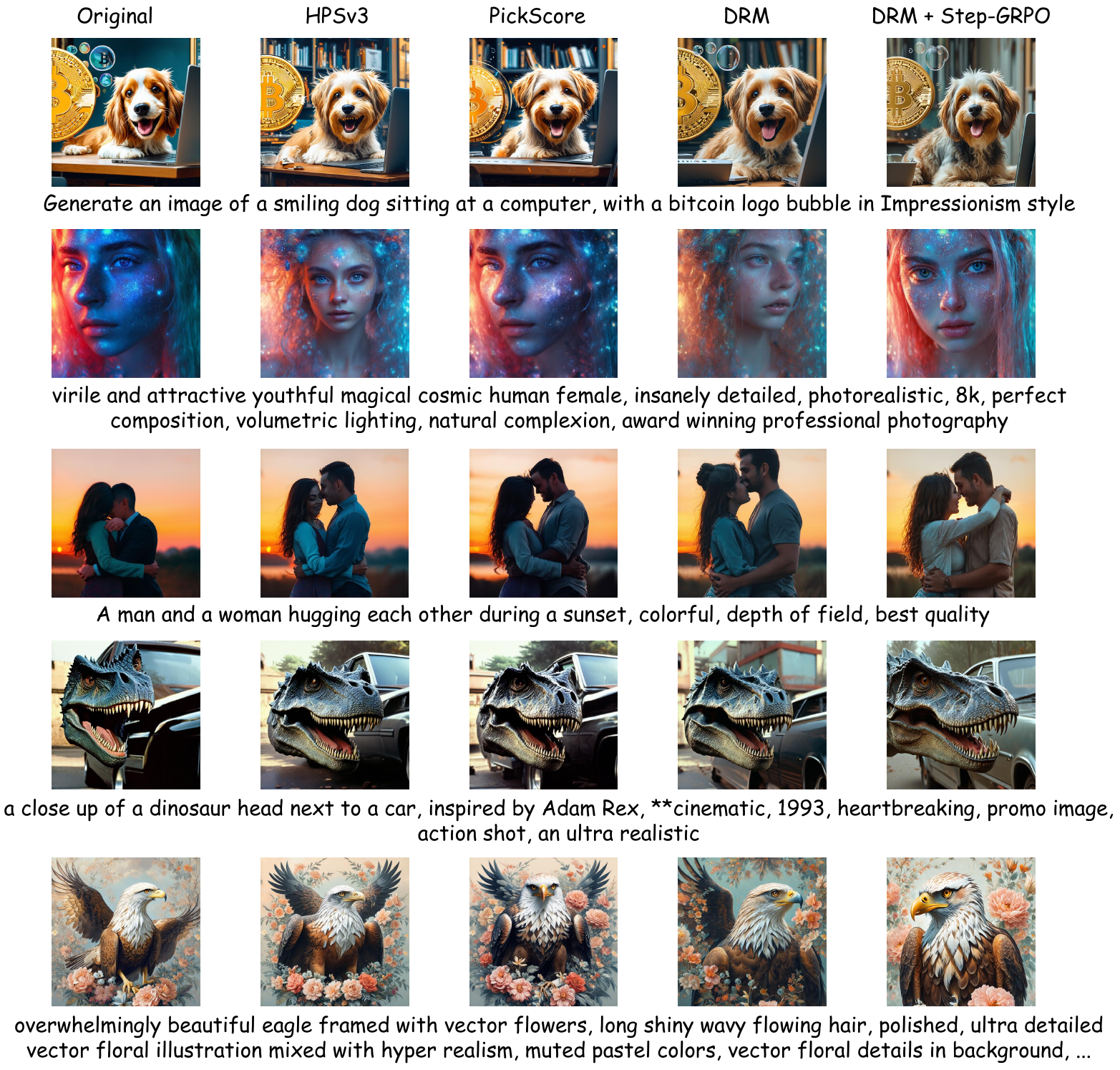}
\vspace{-2.5pt}
\caption{
\textbf{Qualitative comparison of SD3.5-Medium optimized by various reward models. }
Our approach clearly exhibits superior visual quality compared to the competing methods.
}
\label{fig:compare}
\vspace{-15pt}
\end{figure*}

\subsection{Step-Wise GRPO}
\label{sec:stepgrpo}
\subsubsection{Experimental Setting}
\paragraph{Implementation Details.}
In our experiments, we benchmark the effectiveness of three distinct RMs: PickScore and HPSv3, and our proposed DRM.
The base generative model for all experiments is SD3.5-Medium.
To ensure a fair comparison, we fine-tune the generator using the state-of-the-art Flow-GRPO algorithm for all three RMs. 
For our DRM, we also conduct experiments with our novel Step-GRPO algorithm to showcase its unique step-wise guidance capabilities.
For efficient fine-tuning, we optimize the generator using Low-Rank Adaptation (LoRA), with the rank set to 32 and the scaling factor $\alpha$ set to 64.
We use a learning rate of 1e-4 with a policy clipping range of 1e-4.
To ensure a fair comparison, we maintain the same workload across methods: standard GRPO aggregates 6 samples per GPU across 4 GPUs, resulting in a group size of 24 ($6 \times 4$). In contrast, Step-GRPO (default $k=6$) processes the same 24 samples total (6 per GPU) but computes updates using a local group size of 6 without cross-GPU aggregation.
During inference, we employ the Flow Match Euler Discrete Scheduler with 50 sampling steps and a classifier-free guidance (CFG) scale of 4.5.
To maintain a standardized benchmark, the evaluation is conducted on test datasets consistent with the Flow-GRPO.

\paragraph{Evaluation Metrics.}
For a comprehensive and objective evaluation of our method, we employ a suite of automated metrics.
Specifically, we utilize three models as evaluators: PickScore, HPSv3, and ImageReward.
These models are established benchmarks for assessing critical aspects of generation quality, including text-image alignment, aesthetic appeal, and alignment with human preferences.

\begin{table}
  \centering
  \footnotesize
  \setlength\tabcolsep{1.25mm}{
    \resizebox{0.975\linewidth}{!}{
  \scalebox{0.75}{
  \begin{tabular}{lccc}
    \toprule
    \textbf{Model} & \textbf{ImageReward} & \textbf{PickScore} & \textbf{HPSv3} \\
    \midrule
    SD3.5-Medium & 1.01 & 16.76  & 8.95 \\
    + PickScore \& GRPO & 1.14 & 16.94 & 9.64 \\
    + HPSv3 \& GRPO & \underline{1.15} & 16.90 & 9.71 \\
    + DRM \& GRPO & 1.14 & \underline{16.95} & \underline{10.07} \\ 
    \rowcolor{gray!12.5}
    + DRM \& Step-GRPO & \textbf{1.17} & \textbf{17.04} & \textbf{10.28} \\ 
    \bottomrule
  \end{tabular}
  }}}
  \vspace{-1mm}
  \caption{
  \textbf{Performance of SD3.5-Medium on the test set, optimized by different reward models.} Best and second-best results are in \textbf{bold} and \underline{underlined}. Our full approach (DRM \& Step-GRPO) outperforms all baselines, while DRM alone achieves the second-best performance on PickScore and HPSv3, validating the efficacy of both components. }
  \label{table:stepgrpo}
  \vspace{-5mm}
\end{table}

\subsubsection{Quantitative Comparison}
The quantitative results of RL fine-tuning experiments are summarized in Table~\ref{table:stepgrpo}.
We evaluate the alignment of the fine-tuned models with human preferences using three established automated metrics: ImageReward, PickScore, and HPSv3.
As shown, the baseline SD3.5-Medium model serves as our starting point. Applying the standard GRPO algorithm with any of the reward models—PickScore, HPSv3, or our DRM—yields consistent improvements across all evaluation metrics, validating the general effectiveness of RL-based fine-tuning.
Notably, even when constrained to the standard GRPO framework, our DRM demonstrates highly competitive performance, particularly on the HPSv3 metric (10.07).
However, the full potential of our approach is unlocked when our DRM is paired with the Step-GRPO algorithm.
This combination decisively outperforms all other methods, establishing a new state-of-the-art across all three benchmarks.
Specifically, our method achieves top scores of 1.17 on ImageReward, 17.04 on PickScore, and an impressive 10.28 on HPSv3.
This consistent and superior performance provides strong empirical evidence for our central hypothesis: by leveraging a reward model capable of evaluating intermediate generation steps with an algorithm designed to utilize this granular feedback, we can achieve a more effective and robust alignment with human preferences than methods that only provide a final reward.

\subsubsection{Qualitative Comparison}
To complement our quantitative findings, we conduct a qualitative analysis to visually assess the performance of our method.
As illustrated in Figure ~\ref{fig:compare}, images generated by our DRM + Step-GRPO approach exhibit a clear superiority over those from the competing methods.
Specifically, our method renders significantly more fine-grained details and shows a marked reduction in visual artifacts and generation errors.
This advantage is particularly evident in its ability to preserve complex structures and generate realistic textures, areas where other methods often falter.
These qualitative improvements provide compelling visual evidence that our approach excels at generating high-fidelity and aesthetically pleasing images, underscoring the benefits of leveraging step-wise guidance.

\begin{figure}[t]
  \centering
  \includegraphics[width=1.0\linewidth]{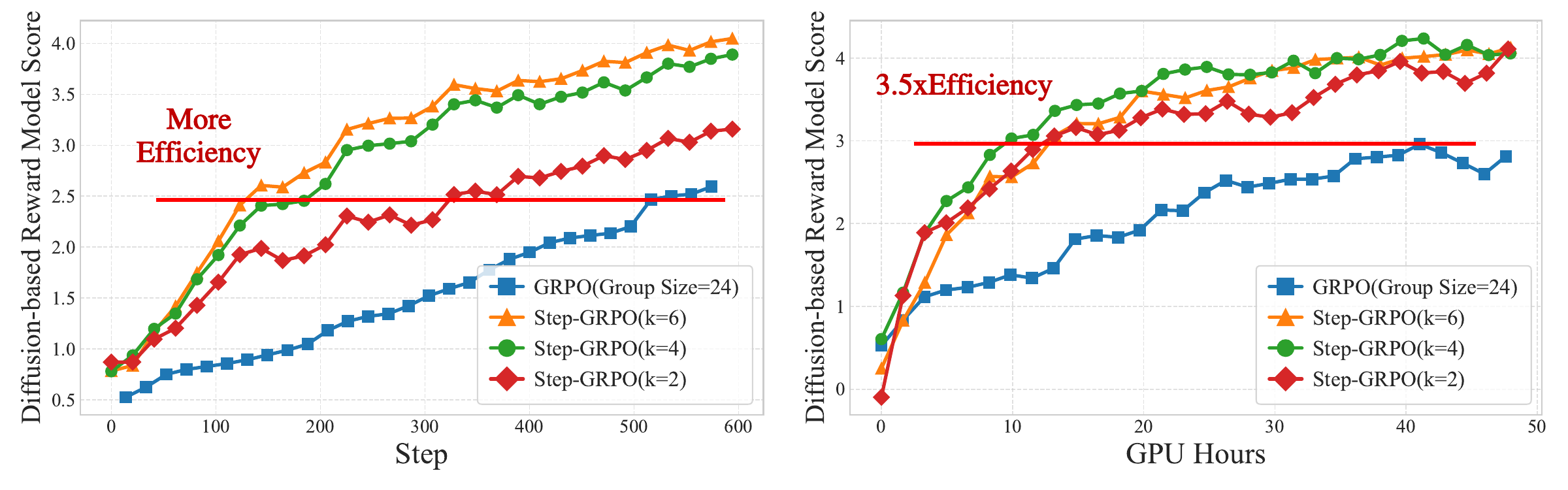}
 \vspace{-7.5pt}
   \caption{Reward curves with steps and GPU hours as the x-axis.}
   \label{fig:reward_line_ablation}
   \vspace{-12.5pt}
\end{figure}

\subsubsection{Training Efficiency and Convergence}
As shown in Figure~\ref{fig:reward_line}, Step-GRPO significantly outperforms standard GRPO in both convergence speed and final performance by leveraging step-wise feedback from a DRM.
We conduct an ablation study on the group size, $k$, to further analyze its properties (Figure~\ref{fig:reward_line_ablation}).
The standard GRPO aggregates 6 samples per GPU across 4 GPUs, resulting in a group size of 24 ($6$$\times$$4$), Step-GRPO ($k$$=$$6$) maintains the same workload (24 samples total, 6 per GPU), but computes updates using a local group size of 6 without cross-GPU aggregation.
For smaller $k$ values, we correspondingly set the per-GPU sample count to $k$ (e.g., 2 samples per GPU for $k$=2).
When measured by steps, our method exhibits superior convergence over GRPO even with $k$$=$$2$, and achieves faster reward growth as $k$ increases (Figure~\ref{fig:reward_line_ablation} (left)).
Regarding \textbf{GPU Hours}, our method converges $\sim$$3.5$$\times$ faster than GRPO (Figure~\ref{fig:reward_line_ablation} (right)).
Notably, smaller $k$ reduces per-iteration computational cost, resulting in similar GPU hour trajectories across $k$$\in$$\{2, 4, 6\}$.


\begin{table}
\centering
\resizebox{\linewidth}{!}{
\begin{tabular}{cccccc}
\toprule
\textbf{Sampling} & \textbf{T(second)$\downarrow$} & \textbf{ImageReward$\uparrow$} & \textbf{PickScore$\uparrow$} & \textbf{HPSv3$\uparrow$} & \textbf{LPIPS$\uparrow$} \\
\hline
$k$=1 & \textbf{2.88} & 1.01 & 16.76 & 8.95 & 0.650 \\
$k$=2 & \underline{5.63} & 1.08 & \underline{16.84} & 9.02 & 0.661 \\
$k$=4 & 7.75 & \underline{1.14} & 16.81 & \underline{9.32} & \textbf{0.663} \\
$k$=6 & 9.83 & \textbf{1.15} & \textbf{16.93} & \textbf{9.49} & \underline{0.662} \\
\bottomrule
\end{tabular}}
\caption{\textbf{Performance of SD3.5-Medium on the test set with and without Step-wise Sampling.} It is evident that applying Step-wise Sampling leads to significant performance gains across all evaluation metrics.}
\label{table:stepsamlping}
\vspace{-5pt}
\end{table}

\begin{figure}[t]
  \centering
    \includegraphics[width=0.875\linewidth]{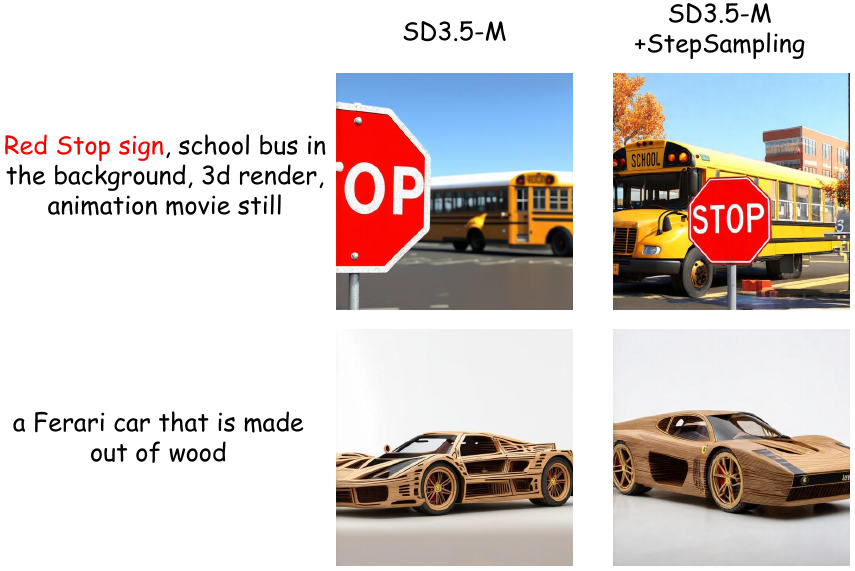}
    \caption{Step-wise Sampling enhances both the fidelity to the prompt and the aesthetic quality of the generated images.}
    \vspace{-10pt}
\label{fig:stepsampling}
\end{figure}

\subsection{Step-wise Sampling}
In addition to Step-GRPO, we investigate the efficacy of Step-wise Sampling. 
Evaluated under the identical protocol described in Section~\ref{sec:stepgrpo}, we further investigate the quality-efficiency trade-off by varying the candidate counts $k \in \{1,2,4,6\}$. 
As presented in Table~\ref{table:stepsamlping}, although generation time (512×512, 50 steps, bfloat16) inevitably scales with $k$, we observe consistent and notable improvements across all human preference metrics.
Additionally, LPIPS evaluations confirm that this approach enhances diversity without inducing mode collapse. 
These quantitative gains are strongly corroborated by qualitative visual comparisons (Figure~\ref{fig:stepsampling}), which reveal superior visual quality, more coherent layouts, and enhanced aesthetic appeal.
Together, these findings validate Step-wise Sampling as an effective inference technique that successfully balances computational cost with elevated generation quality and diversity.


\section{Conclusion}
In this paper, we introduced the Diffusion-based Reward Model (DRM), a novel paradigm that repurposes a pre-trained diffusion model's profound understanding of visual aesthetics as the evaluative backbone.
The DRM's unique ability to assess noisy intermediate latents enabled two key innovations.
For optimization, our Step-wise GRPO leverages dense, per-step rewards to resolve the credit assignment problem, achieving more stable and efficient alignment.
For inference, our Step-wise Sampling strategy uses the DRM as a dynamic guide to proactively steer generation towards higher-quality results.
Our extensive experiments confirm that DRM provides a more powerful solution for aligning diffusion models with human preference.
We hope our work will inspire further exploration for reward modeling.
{
    \small
    \bibliographystyle{ieeenat_fullname}
    \bibliography{main}
}

\clearpage
\maketitlesupplementary
\setcounter{page}{1}
\setcounter{figure}{0}
\setcounter{section}{0}
\renewcommand{\thefigure}{S\arabic{figure}}
\setcounter{table}{0}
\renewcommand{\thetable}{S\arabic{table}}

\begin{table*}[t!]
    \centering
    \resizebox{\linewidth}{!}{
    \begin{tabular}{l|c|cccccc}
    \toprule
    \textbf{Model} & \textbf{Overall $\uparrow$ } & \textbf{Single Obj. $\uparrow$} & \textbf{Two Obj. $\uparrow$} & \textbf{Counting $\uparrow$ } & \textbf{Colors $\uparrow$} & \textbf{Position $\uparrow$} & \textbf{Attr. Binding $\uparrow$} \\
    \midrule
    \multicolumn{8}{c}{\textit{Flow Matching Models}} \\
    \midrule
    FLUX.1 Dev & 0.66 & 0.98 & 0.81 & 0.74 & 0.79 & 0.22 & 0.45 \\
    SD3.5-L & 0.71 & 0.98 & 0.89 & 0.73 & 0.83 & 0.34 & 0.47 \\
    SD3.5-M & 0.63 & 0.98 & 0.78 & 0.50 & 0.81 & 0.24 & 0.52 \\
    \midrule
    \multicolumn{8}{c}{\textit{GRPO based Methods}} \\
    \midrule
    \rowcolor{gray!12.5}
    \textbf{SD3.5-M+Step-GRPO} & \textbf{0.78} & \textbf{0.99} & \textbf{0.93} & \textbf{0.80} & \textbf{0.86} & \textbf{0.37} & \textbf{0.70}  \\
    \bottomrule
    \end{tabular}
    }
    \caption{{\bf GenEval Result.} Results for models are from Flow-GRPO. Obj.: Object; Attr.: Attribution.}
    \label{tab:supp_geneval}
\vspace{-0.5cm}
\end{table*}

\begin{figure}[t]
  \centering
    \includegraphics[width=0.70\linewidth]{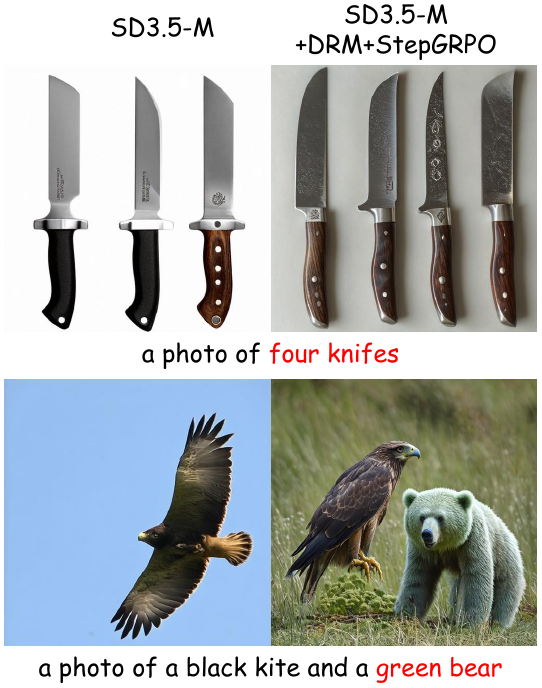}
    \caption{\textbf{Qualitative comparison of generation results before and after optimization with our method.}
    Our approach enhances the model's ability to adhere to complex prompt specifications. 
    (Top) Given a prompt requiring four knives, the baseline model fails on object counting, whereas our optimized model correctly generates the specified number.
    (Bottom) For a prompt requiring specific attribute binding ("a green bear"), the baseline struggles with color assignment, while our model successfully renders the object with the correct attribute.}
    \vspace{0pt}
\label{fig:supp-geneval}
\end{figure}

\section{Result of GenEval}
\label{sec:geneval}
To quantitatively assess text-image alignment, we evaluate our method on the GenEval benchmark.
This benchmark contains 553 prompts designed to test compositional understanding, including object counting, spatial relationships, and attribute binding.
We apply our Step-GRPO to the SD3.5-M model and compare it against several strong flow matching baselines.
Notably, while our reward model is trained exclusively on human preference data, it leads to substantial gains on this objective benchmark. 
As detailed in Table~\ref{tab:supp_geneval}, our method boosts the overall score of SD3.5-M from 0.63 to 0.78, outperforming even the larger SD3.5-L model.
The improvements are particularly pronounced in challenging compositional categories.
For instance, we observe a massive 60\% relative improvement in counting accuracy (from 0.50 to 0.80) and a 35\% relative improvement in attribute binding (from 0.52 to 0.70).
This highlights our method's ability to generalize from subjective preferences to objective prompt fidelity, instilling a more robust understanding of prompt semantics.
These quantitative improvements are further illustrated by the qualitative examples in Figure~\ref{fig:supp-geneval}, where our method successfully handles complex prompts that cause the baseline to fail.

\section{More Visualization Result}
\label{sec:supp_show}
\begin{figure*}[!t]
\vspace{0pt}
\centering
\includegraphics[width=1.0\textwidth]{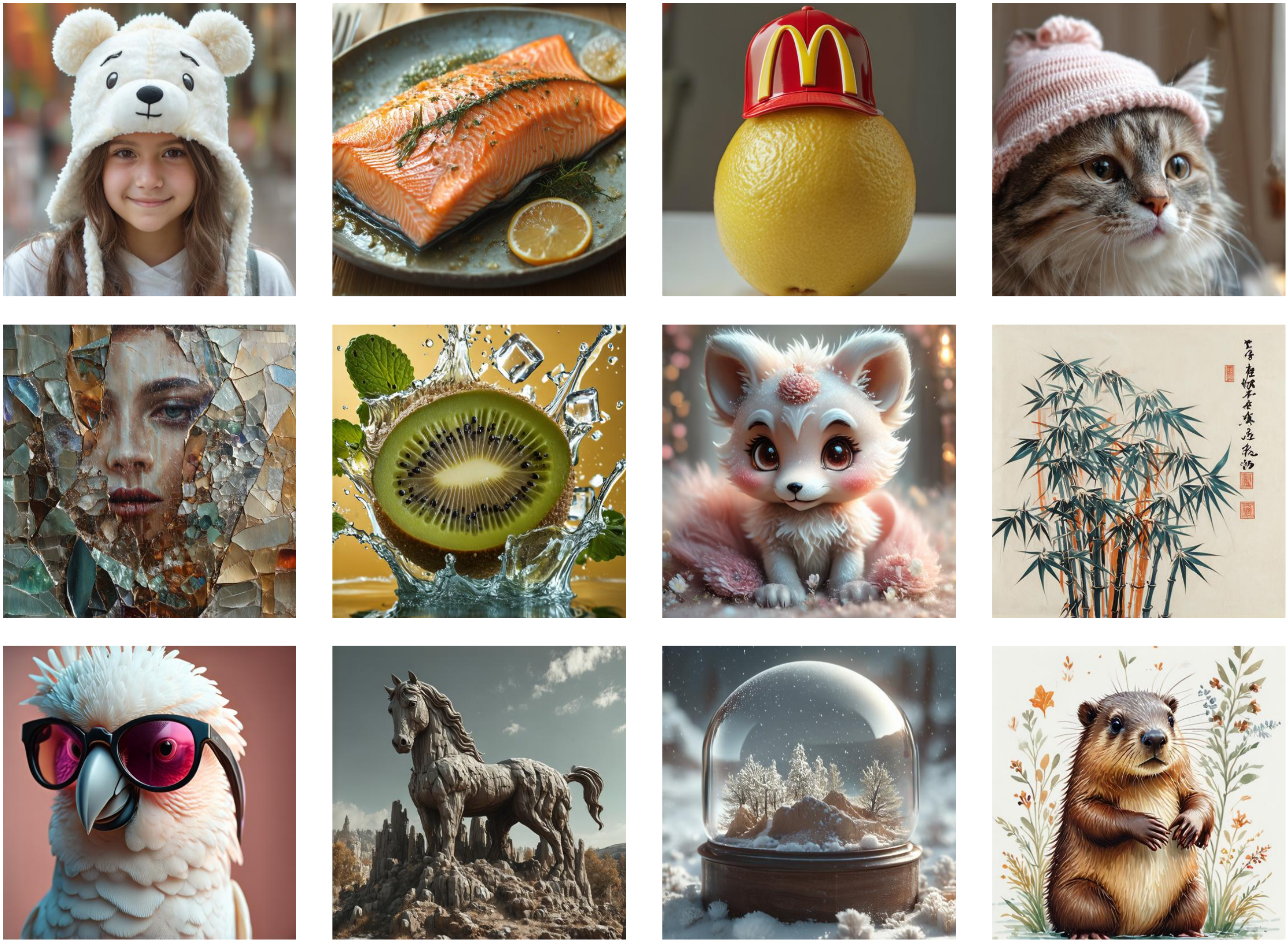}
\vspace{0pt}
\caption{\textbf{Additional qualitative results from our Step-GRPO optimized SD3.5-M model.} These examples showcase the model's ability to generate high-quality, diverse images that faithfully adhere to complex prompts, demonstrating the broad applicability and effectiveness of our method.}
\label{fig:supp-show}
\vspace{0mm}
\end{figure*}

The prompts in Figure~\ref{fig:supp-show} are as follows:
\begin{tcolorbox}[colback=white]
1. 16-year-old teenager wearing a white bear-ear hat with a smirk on their face. \\
2. photo of well done salmon dinner, 8K, Global Illumination, Ray Tracing Reflections \\
3. A lemon with a McDonald's hat. \\
4. cat, cute, hat \\
5. The image is a mixed media collage with broken glass and torn paper elements, featuring intricate oil details and a canvas texture, in a contemporary art style. \\
6. Kiwi fruit, mint leaves, ice cubes, background yellow, splashing water, soft box, back light, creative food photography, Art by Alberto Seveso,
\\
7. little tiny cub beautiful light color White fox soft fur kawaii chibi Walt Disney style, beautiful smiley face and beautiful eyes sweet and smiling features, snuggled in its soft and soft pastel pink cover, magical light background, style Thomas kinkade Nadja Baxter Anne Stokes Nancy Noel realistic \\
8. 185764, ink art, Calligraphy, bamboo plant :: orange, teal, white, black –ar 2:3 –uplight \\
9. A 3D Rendering of a cockatoo wearing sunglasses. The sunglasses have a deep black frame with bright pink lenses. Fashion photography, volumetric lighting, CG rendering. \\
10. A rock formation in the shape of a horse, insanely detailed \\
11. a desert in a snowglobe, 4k, octane render :: cinematic –ar 2048:858 \\
12. watercolour beaver with tale, white background 
\end{tcolorbox}

\end{document}